\pdfoutput=1
\documentclass[11pt]{article}

\usepackage{EACL2023}

\usepackage{times}
\usepackage{latexsym}

\usepackage[T1]{fontenc}

\usepackage[utf8]{inputenc}

\usepackage{microtype}

\usepackage{inconsolata}

\usepackage{paralist, tabularx}

\usepackage{color}

\usepackage{booktabs}       
\usepackage{amsmath}
\usepackage{stmaryrd}
\usepackage{tikz}
\usepackage{multirow}
\usepackage{paralist}
\usetikzlibrary{calc, backgrounds, positioning}
\usepackage{algorithm}
\usepackage{algpseudocode}
\title{Active PETs: Active Data Annotation Prioritisation for Few-Shot Claim Verification with Pattern Exploiting Training}

\author{Xia Zeng, Arkaitz Zubiaga \\
  Queen Mary University of London \\
  \texttt{\{x.zeng,a.zubiaga\}@qmul.ac.uk} \\
}

\begin{document}
\maketitle
\begin{abstract}
To mitigate the impact of the scarcity of labelled data on fact-checking systems, we focus on few-shot claim verification. Despite recent work on few-shot classification by proposing advanced language models, there is a dearth of research in data annotation prioritisation that improves the selection of the few shots to be labelled for optimal model performance. We propose Active PETs, a novel weighted approach that utilises an ensemble of Pattern Exploiting Training (PET) models based on various language models, to actively select unlabelled data as candidates for annotation. Using Active PETs for few-shot data selection shows consistent improvement over the baseline methods, on two technical fact-checking datasets and using six different pretrained language models. We show further improvement with Active PETs-o, which further integrates an oversampling strategy. Our approach enables effective selection of instances to be labelled where unlabelled data is abundant but resources for labelling are limited, leading to consistently improved few-shot claim verification performance.\footnote{Our code is available.}

\end{abstract}

\section{Introduction}

As a means to mitigate online misinformation, research in automated fact-checking has experienced a recent surge of interest. Research efforts have resulted in survey papers covering different perspectives \cite{thorne_automated_2018, kotonya_explainable_2020-1, nakov_automated_2021, zeng_automated_2021, guo_survey_2021} and novel datasets with enriched features \cite{augenstein_multifc_2019, chen_seeing_2019, ostrowski_multi-hop_2020,  jiangHoVerDatasetManyHop2020, schuster_get_2021, aly_feverous_2021, saakyan_covid-fact_2021}. 
Recent work has addressed various challenges, e.g. generating and utilising synthetic data \cite{atanasovaGeneratingLabelCohesive2020, pan_zero-shot_2021, hatuaClaimVerificationUsing2021}, joint verification over text and tables \cite{schlichtkrullJointVerificationReranking2021, kotonyaGraphReasoningContextAware2021}, investigating domain adaptation \cite{liuAdaptingOpenDomain2020a,mithun_data_2021}, achieving better evidence representations and selections \cite{maSentenceLevelEvidenceEmbedding2019, samarinas_improving_2021, siTopicAwareEvidenceReasoning2021, bekoulis_understanding_2021},  and performing subtasks jointly \cite{yinTwoWingOSTwoWingOptimization2018, jiangExploringListwiseEvidence2021a,  zhangAbstractRationaleStance2021a}.

\begin{figure}[t]
    \centering
    \includegraphics[scale=0.4]{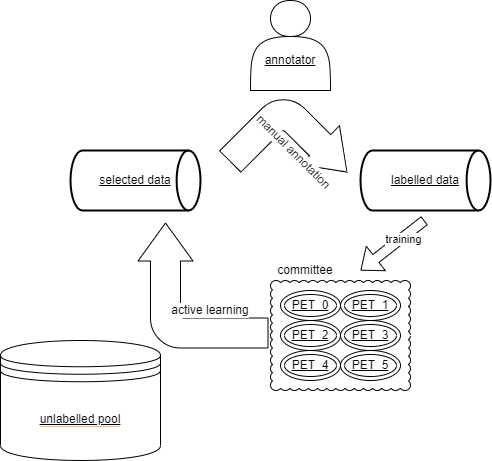}
    \caption{Illustration of the data annotation prioritisation scenario with a committee of 6 PETs.}
    \label{fig:task}
\end{figure}

As a core component of a fact-checking system, a claim validation pipeline consists of document retrieval, rationale selection and claim verification \cite{zeng_automated_2021}. Our main focus here is claim verification, the task of assessing claim veracity with retrieved evidence. It is typically treated as a natural language inference (NLI) task: given a claim and an evidence, the aim is to predict the correct veracity label out of ``Support'', ``Neutral'' and ``Contradict''.
Substantial improvements have been achieved in the performance of claim validation models when a considerable amount of training data is available \cite{pradeep_scientific_2020, li_paragraph-level_2021, zeng_qmul-sds_2021, zhang_abstract_2021, wadden_longchecker_2021}. However, where new domains needing fact-checking emerge, collecting and annotating new relevant datasets can carry an impractical delay. Availability of unlabelled data can often be abundant, but given the cost and effort of labelling this data, one needs to be selective in labelling a small subset. In these circumstances, rather than randomly sampling this subset, we propose to optimise the selection of candidate instances to be labelled through active learning, such that it leads to overall improved few-shot performance.

To the best of our knowledge, our work represents the first such effort in proposing an approach leveraging an active learning strategy for the claim verification problem, as well as the first in furthering Pattern Exploiting Training (PET) with an active learning strategy. To achieve this, we propose Active PETs, a novel methodology that enables the ability to leverage an active learning strategy through a committee of PETs. Figure \ref{fig:task} illustrates the application of the active learning strategy on claim verification.


By exploring effective prioritisation of unlabelled data for annotation and making better use of a small amount of labelled data, we make the following novel contributions:

\begin{compactitem}
    \item we are the first to study data annotation prioritisation through active learning for few-shot claim verification;
    \item we are the first to study the extensibility of PET to enable active learning, by proposing Active PETs, a novel ensemble-based cold-start active learning strategy that enables multiple pretrained language models (PLMs) to collectively prioritise data instances;
    \item we further investigate the effect of oversampling on mitigating the impact of imbalanced data selection on few-shot learning, when guided by active learning;
    \item we conduct further corpus-based analysis on the selected few-shot data instances, which highlights the potential of Active PETs to lead to improved lexical and semantic characteristics that benefit the task.
\end{compactitem}

Our results show consistently improved performance of Active PETs over the baseline active learning strategies on two datasets, SCIFACT \cite{wadden_fact_2020} and  Climate FEVER \cite{diggelmann_climate-fever_2021}. In addition to improved performance over the baselines, our research emphasises the importance of the hitherto unexplored data prioritisation in claim verification, showing remarkable performance improvements where time and budget are limited.

\section{Background}

\subsection{Claim Verification}

Claim verification is typically addressed as an NLI problem \cite{thorne_automated_2018}. Recent progress has enforced a closed-world reliance \cite{pratapaConstrainedFactVerification2020} and incorporated multiple instance learning \cite{satheAutomaticFactCheckingDocumentlevel2021}. While data scarcity poses a major challenge on automated fact-checking \cite{zeng_automated_2021}, research on few-shot claim verification is limited to date. \citet{lee_towards_2021} investigated a perplexity-based approach that solely relies on perplexity scores from PLMs. Their model was tested on binary claim verification, as opposed to the three-way classification in our work. \citet{zengAggregatingPairwiseSemantic2022} introduced SEED, a vector-based method that aggregates pairwise semantic differences for claim-evidence pairs to address the task of few-shot claim verification. While their model addresses three-way classification, the experiments are only conducted in ideal scenarios where oracle evidences are available. To the best of our knowledge, however, no work has investigated the use of active learning in the context of claim verification. To further research in this direction, we propose Active PETs, a model that incorporates active learning capabilities into PET \cite{schick_exploiting_2021, schick_its_2021}. PET has shown competitive performance in a range of NLP classification tasks, but its adaptation to the context of automated fact-checking and/or active learning settings has not been studied.

\subsection{Active Learning}

Active Learning (AL) is a paradigm used where labelled data is scarce \cite{ein-dor_active_2020}. The key idea is that a strategic selection of training instances to be labelled can lead to improved performance with less training \cite{settles_active_2009}. Active learning methods are provided with an unlabelled pool of data, on which a querying step is used to select candidate instances to be annotated with the aim of optimising performance of a model trained on that data. The goal is therefore to optimise performance with as little annotation --and consequently budget-- as possible. Traditional active learning query strategies mainly include uncertainty sampling, query-by-committee (QBC) strategy, error/variance reduction strategy and density weighted methods \cite{settles_active_2012}. Recent empirical studies have revisited the traditional strategies in the context of PLMs: \citet{ein-dor_active_2020} examined various active learning strategies with BERT \cite{devlin_bert_2019}, though limited to binary classification tasks. \citet{schroderRevisitingUncertaintybasedQuery2022} conducted experiments with ELECTRA \cite{clark_electra_2020}, BERT, and DistilRoBERTa \cite{sanh_distilbert_2020} respectively, while limiting the scope to uncertainty-based sampling.

Recent efforts on combining active learning with PLMs go into both warm-start and cold-start strategies. Warm-start strategies require a small initial set of labelled data to select additional instances, while cold-start strategies can be used without an initial set of labelled data. \citet{ash_deep_2020} proposed Batch Active learning by Diverse Gradient Embeddings (BADGE) that samples a batch of instances based on diversity in gradient loss. \citet{margatina_active_2021} proposed Contrastive Active Learning (CAL), the state-of-the-art (SOTA) warm-start strategy that highlights data with similar feature space but maximally different predictions. Furthermore, Active Learning by Processing Surprisal (ALPS) \cite{yuan_cold-start_2020}, the SOTA cold-start strategy, utilises masked language model (MLM) loss as an indicator of model uncertainty. We use BADGE, CAL and ALPS for baseline comparison, please see detailed descriptions in section \ref{baselines}. 

To the best of our knowledge, QBC strategies \cite{seung_query_1992, dagan_committee-based_1995, freund_selective_1997} that utilise a committee of models remains to be explored with PLMs, as previous studies limit their scope at measuring single model uncertainty. Nowadays various PLMs are publicly available that applying an ensemble-based query strategy on a downstream task becomes realistic, especially in few-shot settings where the computation required is relatively cheap. Furthermore, previous studies always perform fine-tuning to get classification results from PLMs. Our work presents the first attempt at integrating an active learning strategy into PET, which we investigate in the context of claim verification for fact-checking.





\section{Methodology}

In this section, we first describe PET, then introduce our model Active PETs, and finally describe the oversampling mechanism we use.

\subsection{Pattern Exploiting Training}

\begin{figure}[t]
    \centering
    \includegraphics[scale=0.55]{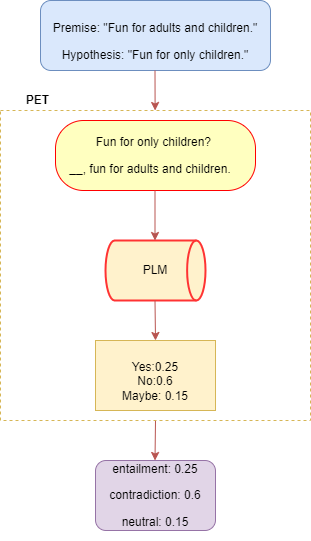}
    \caption{An example of doing NLI with PET.}
    \label{fig:PET}
\end{figure}

Pattern Exploiting Training (PET) \cite{schick_exploiting_2021, schick_its_2021} is a semi-supervised training procedure that can reformulate various classification tasks into cloze questions with natural language patterns and has demonstrated competitive performance in various few-shot classification tasks. To predict the label for a given instance $x$, it is first reformulated into manually designed patterns that have the placeholder $[mask]$. Then, the probability of each candidate token for replacing $[mask]$ is calculated by using a pretrained language model, where each candidate is mapped to a label according to a manually designed verbaliser.
Figure \ref{fig:PET} provides an example of NLI using PET. 

\subsection{Proposed method: Active PETs}

Having a large pool of unlabelled data, our objective is to design a query strategy that selects suitable candidates to be labelled, such that the labelled pool of instances leads to optimal few-shot performance. Our query strategy is rooted in the intuition that disagreement among different PETs in a committee can capture the uncertainty of a particular instance.

Based on the assumption that performance of different language models is largely dependent on model size \cite{kaplan_scaling_2020}, we introduce a weighting mechanism: each PET is first assigned a number of votes $V_i$ that is proportional to its hidden size,\footnote{For example, if we use a committee formed of only base models that have 6 hidden layers and large models that have 12 hidden layers, proportionally each of the base models is allocated one vote and each of the large models is allocated two votes.} and ultimately all votes are aggregated (see Algorithm \ref{query} in appendix \ref{algo}). 

We then quantify the disagreement by calculating vote entropy \cite{dagan_committee-based_1995}:

\begin{equation}
    score_x = -\sum_{\hat{y}}\frac{vote(x,\hat{y})}{count(V)} \log{\frac{vote(x,\hat{y})}{count(V)}}
\end{equation}

where $\hat{y}$ is the predicted label, $x$ is the instance, $vote(x,\hat{y})$ are the committee votes of $\hat{y}$ for the instance $x$, and $count(V)$ is the number of total assigned votes. It can be viewed as a QBC generalisation of entropy-based uncertainty sampling that is designed to combine models of different sizes.

\subsection{Data Oversampling}
\label{ssec:oversampling}

One of the risks of the proposed active learning strategy is that the resulting training data may not be adequately balanced, which can impact model performance. An accessible solution is oversampling: resample the instances from the minority class with replacement until balanced. Note that this does not increase the labelling effort as instances are repeated from the labelled pool. Instead of random resampling \cite{japkowicz_class_2000}, we propose a novel technique of integrating resampling with the committee's preference. For each minority class, we start resampling from the instance that has the highest disagreement score to the instance that has the lower disagreement score. In highly imbalanced cases, resampling is repeated from the highest to lowest priority until the overall label distribution is balanced (see Algorithm \ref{training} in appendix \ref{algo}).

\section{Experimental Settings}

Here we present the datasets and models used.

\subsection{Datasets}



\begin{table}[htb]
    \footnotesize
    \centering
    \begin{tabular}{cccc}
        \toprule
        \multicolumn{4}{c}{\textbf{SCIFACT}} \\
        \midrule
         & \textbf{`Support'} & \textbf{`Neutral'} & \textbf{`Contradict'} \\
        \midrule
        \textbf{UP} & 266 (9.31\%) & 2530 (88.55\%) & 61 (2.14\%) \\
        \midrule
        \textbf{Test} & 150 (33.33\%) & 150 (33.33\%) & 150 (33.33\%) \\
        \bottomrule
        \toprule
        \multicolumn{4}{c}{\textbf{cFEVER}} \\
        \midrule
         & \textbf{`Support'} & \textbf{`Neutral'} & \textbf{`Contradict'} \\
        \midrule
        \textbf{UP} & 1789 (24.78\%) & 4778 (66.19\%) & 652 (8.66\%) \\
        \midrule
        \textbf{Test} & 150 (33.33\%) & 150 (33.33\%) & 150 (33.33\%) \\
        \bottomrule
    \end{tabular}
    \caption{Label distribution of SCIFACT and cFEVER. UP = unlabelled pool of training data.}
    \label{distribution}
\end{table}

We choose real-world datasets with real claims, SCIFACT and Climate FEVER, known to be challenging, technical and free of synthetic data.\footnote{See data samples in Appendix \ref{sec:appendix}.}

\textbf{SCIFACT} provides scientific claims with their veracity labels, as well as a collection of scientific paper abstracts, some of which contain rationales to resolve the claims. In addition, it provides the oracle rationales that can be linked to each claim.

For SCIFACT, we perform the pipeline including abstract retrieval and claim verification. For the abstract retrieval step, we use BM25 to retrieve the top 3 abstracts, skipping the more specific rationale selection, as the SOTA system for this dataset suggested \cite{wadden_longchecker_2021}. We chose BM25 based on high recall results reported in previous work \cite{pradeep_scientific_2020}. We merge original SCIFACT train set and dev set and redistribute the data to form a test set that contains 150 instances for each class and use the rest in the unlabelled pool. The reformulated data is highly imbalanced as presented in Table \ref{distribution}.


\textbf{Climate FEVER (cFEVER)} is a challenging large-scale dataset that consists of claim and evidence pairs on climate change, along with their veracity labels. As it does not naturally provide options of setting up retrieval modules, we directly use it on the task of claim verification. Similarly we reserve 150 instances for each class to form a test set and leave the rest in the unlabelled pool. Data in the unlabelled pool is heavily skewed, as shown in Table \ref{distribution}.

\subsection{Active PETs}

Committees of five to fifteen models are common for an ensemble-based active learning strategy \cite{settles_active_2012}. Here we form a committee of 6 PETs with 3 types of PLMs: BERT-base, BERT-large \cite{devlin_bert_2019}, RoBERTa-base, RoBERTa-large \cite{liu_roberta_2019}, DeBERTa-base and DeBERTa-large \cite{he_deberta_2021}. Given the commonalities between the NLI and claim verification tasks, we use the PLM checkpoints already fine-tuned on MNLI \cite{williams_broad-coverage_2018}. Note that the committee of 6 PETs is used for data selection, and then we train a PET based on each PLM individually for separate experiments, and hence we present results for 6 different PETs separately.

Despite a line of research in optimising PET patterns and verbalisers \cite{tam_improving_2021}, that is not our main focus. We use the following pattern and verbaliser for PET: \underline{[claim]?\_\_, [evidence]}; ``Support'':``Yes'', ``Contradict'':``No'', ``Neutral'':``Maybe'', as they yielded best performance on NLI tasks in our preliminary experiments.

We test two variants: \textbf{Active\_PETs} with no oversampling, and \textbf{Active\_PETs-o} with the oversampling described in Section \ref{ssec:oversampling}.

\subsection{Baselines}
\label{baselines}

We compare our method to four baselines: random sampling, BADGE, CAL and ALPS.

\subsubsection{Random sampling}
For random sampling, we run each experiment over 10 different sampling seeds ranging from 123 to 132, and present the averaged results.

\subsubsection{BADGE}
BADGE \cite{ash_deep_2020} optimises for both uncertainty and diversity. Gradient embeddings $g_x$ are first computed for each data in the unlabelled pool, where $g_x$ is the gradient of the cross entropy loss with respect to the parameters of the model’s last layer. It then applies k-MEANS++ clustering on the obtained gradient embeddings, and batch selects instances that differ in feature representation and predictive uncertainty.

Though BADGE is proposed as a warm-start method, the required initial set of labelled data is only used for the initial training the model. In our experiments on claim verification, PLMs that are already finetuned on a similar task NLI are used, hence, BADGE can be used for cold-start sampling.

\subsubsection{CAL}
CAL \cite{margatina_active_2021}, the SOTA warm-start strategy, highlights contrastive data points: data that has similar model encodings but different model predictions. Unlike BADGE, an initial labelled set of data is essential for CAL. It first calculates the [CLS] embeddings for all of the data and then runs K-Nearest-Neighbours (KNN) to obtain the \emph{k} closest labelled neighbours for each unlabelled instance. It further calculates predictive probabilities from the model and measures Kullback-Leibler divergence on it. Finally it selects unlabelled instances whose predictive likelihoods diverge the most from their neighbours.

While CAL achieves SOTA performance as a warm-start strategy, its dependence on an initial labelled set of data makes it incompatible in the same few-shot active learning settings without an initial labelled set. However, for comprehensive comparison purposes, we still include it as a baseline starting at 100 labelled instances that are obtained from random sampling with 10 different random seeds.

\subsubsection{ALPS}

ALPS \cite{yuan_cold-start_2020}, the SOTA cold-start active learning method, also aims to take both model uncertainty and data diversity into account. It calculates surprisal embeddings to represent model uncertainty. Specifically, for each instance $x$, it is passed through the masked language modelling head of a PLM and then 15\% of the tokens in $x$ are randomly selected to calculate the cross entropy against their target tokens. The surprisal embeddings go through L2-normalisation and then get clustered to select the top samples.

\begin{figure*}[htb]
    \centering
    \includegraphics[width=16cm,height=3cm]{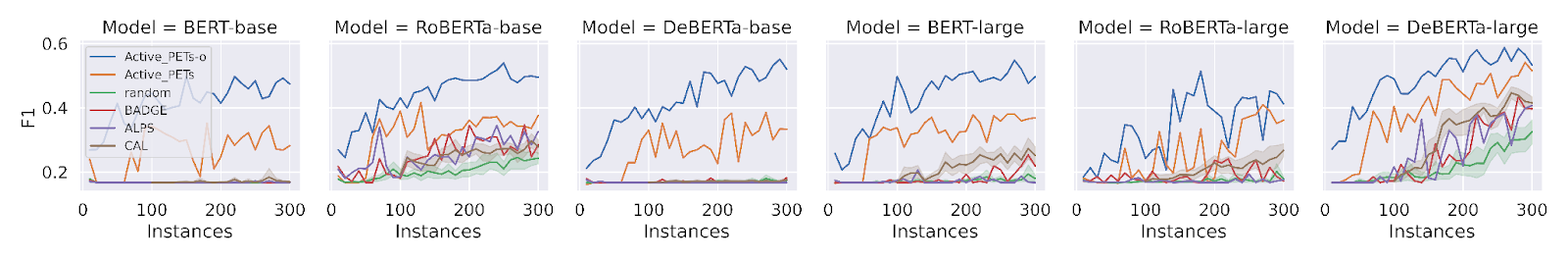}
    \caption{Few-Shot F1 Performance on SCIFACT claim verification.}
    \label{fig:scifact}
\end{figure*}

\subsection{Training Details}
\noindent \textbf{Hyperparameters.} As in few-shot settings we lack a development set, we follow previous work \cite{schick_exploiting_2021,schick_its_2021} and use the following hyperparameters for all experiments: $1e^{-5}$ as learning rate, 16 as batch size, 3 as the number of training epochs, 256 as the max sequence length. \footnote{See further details for reproducibility in Appendix \ref{reproducibility}.}

\noindent \textbf{Labelling budget.} We set it to a maximum of 300. We experiment with all scenarios ranging from 10 to 300 instances with a step size of 10.

\noindent \textbf{Checkpoints.} We always use the PLM checkpoints from the last iteration to perform active learning, but always train the initial PLMs which have never been trained on any fact-checking datasets.

\section{Results}
\begin{figure*}[htb]
    \centering
    \includegraphics[width=16cm,height=3cm]{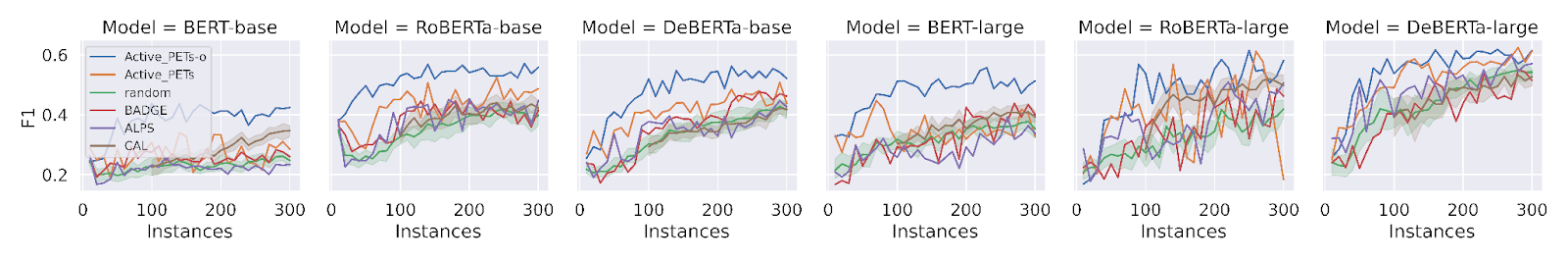}
    \caption{Few-Shot F1 Performance on cFEVER claim verification.}
    \label{fig:cfever}
\end{figure*}

We next discuss results for our experiments.

\subsection{Results on SCIFACT}

Figure \ref{fig:scifact} presents experimental results on SCIFACT, where the unlabelled pool is large, heavily imbalanced and the domain is technical. Each subfigure shows results for a different PET among the six under consideration.

Data retrieved with Active PETs brings substantial improvements for all of the models, often from the very beginning but consistently as the number of shots increases from around 50 instances. Despite the performance fluctuations, training using data sampled with Active PETs rarely underperforms the baselines for SCIFACT. With Active PETs, Bert-base peaks at 0.352, RoBERTa-base peak at 0.345; DeBERTa-base peaks at 0.385; BERT-large peaks at 0.380; RoBERTa-large peaks at 0.409; DeBERTa-large peaks at 0.541. Generally, Active PETs shows a 10 to 20\% increase in F1 scores, compared with various baselines.

Moreover, with Active PETs-o, i.e. when oversampling is further integrated with Active PETs, we observe a significant performance increase. Models tend to learn better from the beginning; the increase trend has less fluctuation; and the overall F1 scores are much higher. In this case, Bert-base peaks at 0.497, RoBERTa-base peak at 0.539; DeBERTa-base peaks at 0.551; BERT-large peaks at 0.548; RoBERTa-large peaks at 0.514; DeBERTa-large peaks at 0.587. This highlights the potential of oversampling, which increases the number of instances without additional labelling budget.

Among the baselines, we observe that training with data retrieved from all baselines failed to lead to any effective outcomes for BERT-base and DeBERTa-base within a labelling budget of 300 instances. While BADGE and CAL lead to some improvements over BERT-large and RoBERTa-large when given over 100 instances, random and ALPS failed to bring any improvements. Baseline results are better with RoBERTa-base and DeBERTa-large, but underperform Active PETs. 


\subsection{Results on cFEVER}

Figure \ref{fig:cfever} presents F1 scores on cFEVER, where the unlabelled pool is large, imbalanced and the domain is somewhat technical. In this case, models generally achieve higher F1 scores than on SCIFACT. First of all, we observe that Active PETs outperforms random baseline in a more stable manner. It is over 10\% higher than random most of the time, although it shows large performance fluctuations on RoBERTa-large. With Active PETs, Bert-base peaks at 0.34, RoBERTa-base peak at 0.524; DeBERTa-base peaks at 0.508; BERT-large peaks at 0.447; RoBERTa-large peaks at 0.612; DeBERTa-large peaks at 0.624. Moreover, Active PETs-o leads to a further performance boost, and more importantly, smooths out the large performance fluctuations. It is about 20\% better than the random baseline most of the time. Specifically, Bert-base peaks at 0.438, RoBERTa-base peak at 0.571; DeBERTa-base peaks at 0.562; BERT-large peaks at 0.557; RoBERTa-large peaks at 0.615; DeBERTa-large peaks at 0.618.

When it comes to the baselines, the baselines do not struggle as much in the worst cases. Even if BERT-base's performance merely increased with most of the baselines, all of the other models managed to improve within the budget. With random sampling, RoBERTa-base, DeBERTa-base, BERT-large and RoBERTa-large all roughly peak at around 0.4, while DeBERTa-large is much better and peaks at around 0.5. BADGE, CAL and ALPS are in general better than random, but achieves lower F1 scores than Active PETs, especially in few-shot settings when the labelling budge is below 100. 


\begin{figure*}[htb]
    \centering
    \includegraphics[width=16cm,height=3cm]{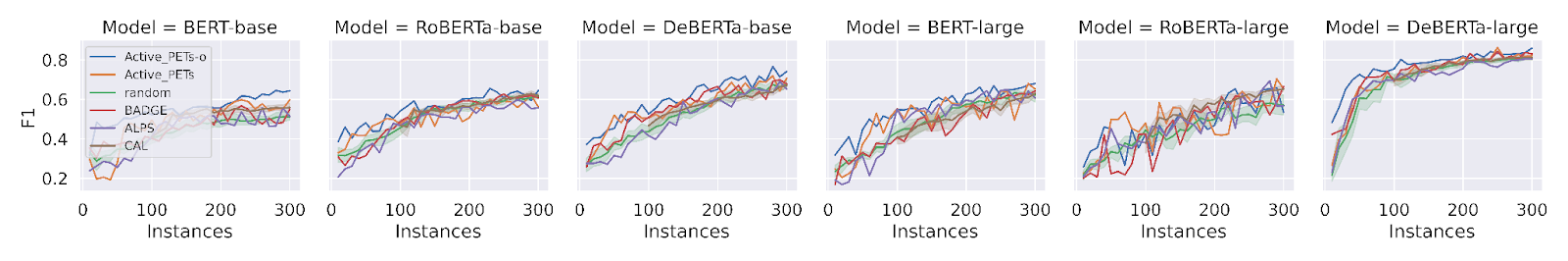}
    \caption{Few-Shot F1 Performance on Oracle SCIFACT claim verification.}
    \label{fig:scifact_oracle}
\end{figure*}

\section{Ablation Study}

With SCIFACT we designed a slightly different pipeline where we conduct both evidence retrieval and claim verification -- a setting that wasn't provided with cFEVER. To assess the impact of the addition of the evidence retrieval component on SCIFACT, we further perform ablation experiments on SCIFACT with oracle evidence. 

With oracle evidence, the number of ``Neutral'' claim-evidence pairs are significantly reduced, resulting in a more balanced overall label distribution. After reserving 100 instances from each class for the test set, the unlabelled pool has 765 instances in total, where ``Support'' takes 46.54\%, ``Neutral'' takes 38.43\% and ``Contradict'' takes 15.03\%. As shown in Figure \ref{fig:scifact_oracle}, overall few-shot performance is much better and active learning demonstrates lesser performance gains. Sampling with baseline active learning strategies in general leads to similar results as random sampling. 
Surprisingly, coupling Active PETs with oversampling when the labelled pool is reasonably balanced, still maintains performance advantages across models. Under this setting, Bert-base peaks at 0.645, RoBERTa-base peak at 0.655; DeBERTa-base peaks at 0.766; BERT-large peaks at 0.68; RoBERTa-large peaks at 0.657; DeBERTa-large peaks at 0.86.

As demonstrated above, active learning is much more helpful for SCIFACT in a real-world setting than in an oracle setting. We could expect that if this finding generalises to cFEVER, active learning in a real-world setting involving evidence retrieval could possibly lead to larger performance gains.

\section{Analysis}

To better understand the impact of data prioritisation, we delve into the labelled data. In the interest of focus, we compare Active PETs with the SOTA cold-start method ALPS by analysing the best-performing PLM DeBERTa-large where 300 instances are selected.

\subsection{Balancing Effects}

\begin{figure}[htb]
    \centering
    \includegraphics[scale=0.35]{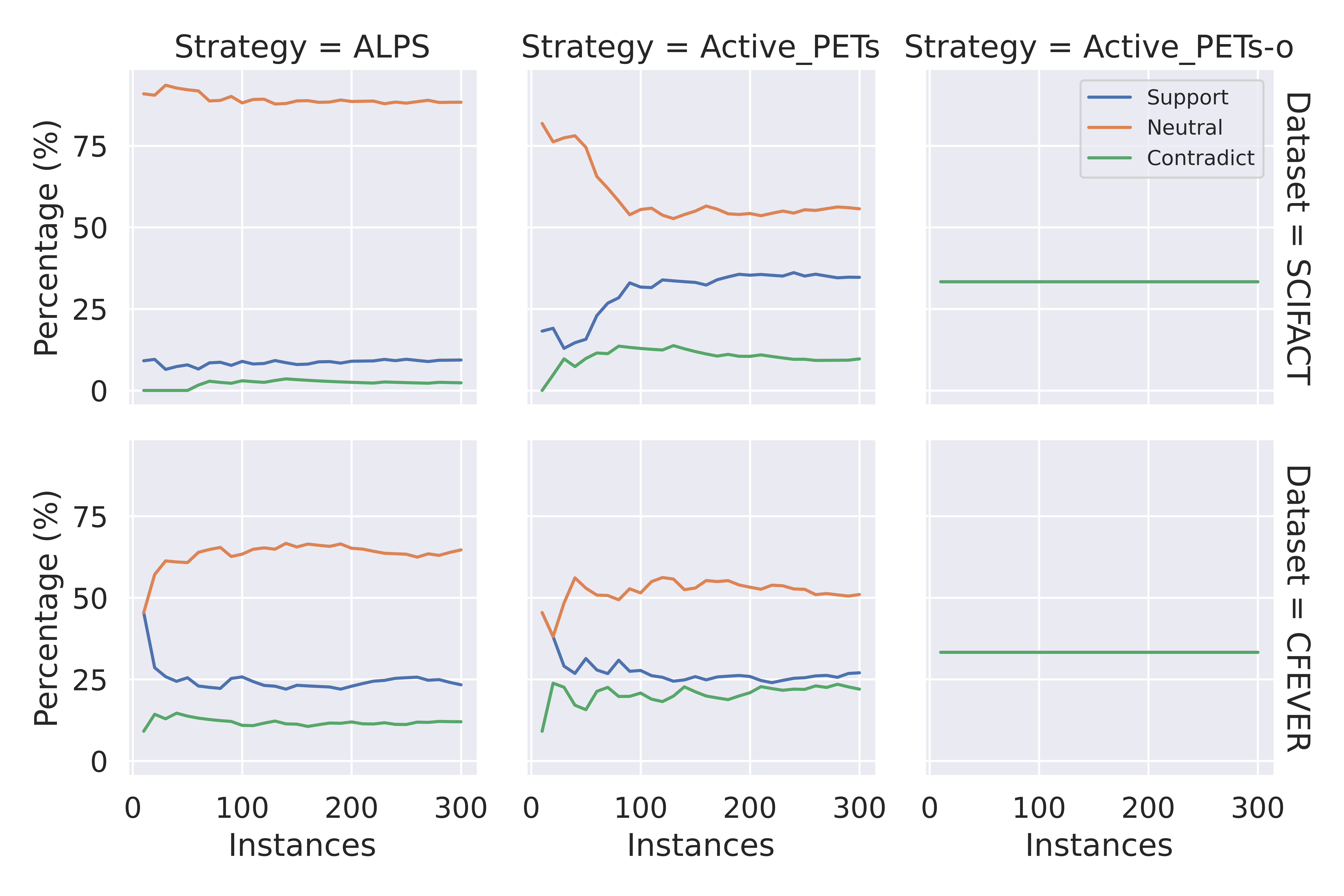}
    \caption{Label Distribution of data obtained with active learning by DeBERTa-large. The upper row is for SCIFACT and the lower row is for cFEVER. 
    }
    \label{fig:class}
\end{figure}

We first look at the distribution of labels for the selected data. Figure \ref{fig:class} shows remarkable difference on label distribution for different active learning strategies. ALPS samples over 80\% data from ``Neutral'', less than 10\% from ``Support'' and very few from ``Contradict'' for SCIFACT; over 60\% data from ``Neutral'', over 20\% from ``Support'' and less than 20\% from ``Contradict'' for cFEVER. They correlate well with original label distribution of each unlabelled pool, as presented in table \ref{distribution}. It suggests that ALPS is not sensitive to label distribution. However, Active PETs manages to sample a much more balanced distribution out of the extremely skewed original distribution. For SCIFACT, despite the initial few iterations, Active PETs samples less than 60\% data from ``Neutral'', less than 40\% data from ``Support'', around 10\% data from ``Contradict''; for cFEVER, Active PETs samples less than 60\% data from ``Neutral'', over 20\% data from ``Support'', around 20\% data from ``Contradict''. In both datasets, label distribution from Active PETs are significantly more balanced than ALPS. Finally, the strategy with oversampling returns perfectly balanced distribution as expected. We identify a strong correlation between label distribution and classification performance.

\subsection{Linguistic Effects}

Aiming at providing further insights into data quality, we conduct corpus-based linguistic analysis to investigate lexical richness and semantic similarity.
\begin{table}[htb]
    \footnotesize
    \centering
    \begin{tabular}{cccc}
        \toprule
        \multicolumn{4}{c}{\textbf{Lexical Richness}} \\
        \midrule
         & \textbf{ALPS} & \textbf{Active\_PETs} & \textbf{Active\_PETs-o} \\
        \midrule
        \textbf{SCIFACT} & 0.0362 & 0.0387 & 0.0447 \\
        \midrule
        \textbf{cFEVER} & 0.0389 & 0.0413 & 0.0503 \\
        \bottomrule
        \toprule
        \multicolumn{4}{c}{\textbf{Semantic Similarity}} \\
        \midrule
         & \textbf{ALPS} & \textbf{Active\_PETs} & \textbf{Active\_PETs-o} \\
        \midrule
        \textbf{SCIFACT} & 0.7921 & 0.8031 & 0.8054 \\
        \midrule
        \textbf{cFEVER} & 0.7449 & 0.7744 & 0.7841 \\
        \bottomrule
    \end{tabular}
    \caption{Lexical richness is measured with Maas Type-Token Ratio (MTTR) scores and Semantic Similarity is measured by cosine similarity scores on embeddings of claims and evidences.}
    \label{analysis}
\end{table}

\subsubsection{Lexical Richness}

A popular metric for calculating lexical richness is Type-Token Ratio (TTR), where the total number of unique tokens is divided by the total number of tokens. We use Maas Type-Token Ratio (Maas TTR) \cite{maas_uber_1972}, a logarithmic variant of TTR, which is demonstrated to be less sensitive to the length of the text \cite{mccarthy_vocd_2007}:

\begin{equation}
    a^2 = \frac{logN -logV}{logN^2}
\end{equation}
where $N$ is the number of tokens in the corpus and $V$ is the number of unique tokens in the corpus.

As shown in the upper part of Table \ref{analysis}, data selected by ALPS has the lowest lexical richness, while Active PETs leads to higher lexical richness for both datasets. Even more surprisingly, when integrating Active PETs with oversampling, the corpus has even higher score at lexical richness, despite that there are multiple duplicated instances in the corpus. One possibility is that training data with higher lexical richness may convey more useful information, as a bigger vocabulary enables more precise expressions.

\subsubsection{Semantic Similarity}

To investigate the overall data diversity, we calculate the average semantic similarity of all possible claim-evidence pairs in the corpus.\footnote{Note that if we only calculate the retrieved pairs, the average similarity scores are approximately 1 for all strategies.} We obtain embeddings of claims and evidences with the PLM at interest, namely DeBERTa-large that has been trained  on MNLI. For each embedded claim, we calculate its cosine similarity score with all embedded evidences in the corpus. The average of all similarity scores is then obtained. The lower part of Table \ref{analysis} shows that ALPS leads to lowest overall semantic embedding similarity scores and Active PETs leads to higher scores. Integrated with oversampling, Active PETs leads to even higher similarity scores. It correlates well with the design of the strategies: ALPS explicitly encourages data diversity, while Active PETs focuses on committee uncertainty. One possible explanation is that data diversity is not as beneficial when the unlabelled pool contains less relevant instances: in the case of SCIFACT and cFEVER datasets, the majority of the unlabelled pool belongs to the ``Neutral'' class where the evidence is not enough to reach a verdict for the claim.

\section{Conclusions}

We present the first study on data annotation prioritisation for claim verification in automated fact-checking. With our novel method Active PETs, we demonstrate the potential of utilising a committee of PETs to collaboratively select unlabelled data for annotation, furthering in turn the extensibility of PET to active learning for the first time. Experiments on the SCIFACT and cFEVER datasets demonstrate the effectiveness of our proposed method, particularly in dealing with imbalanced data. Our proposed model consistently outperforms the random, BADGE, CAL and ALPS baselines by a margin. Further integration with an oversampling strategy that does not impact labelling effort leads to consistent performance improvements in all tested settings. Data that is more balanced shows to have higher lexical richness and semantic similarity, leading to better training results. While we have shown its effectiveness for claim verification here, in the future we aim to investigate Active PETs in other downstream tasks.

\section{Limitations}
We focus on demonstrating the effectiveness of Active PETs in scenarios where the labelling budget is limited and the label distribution is very imbalanced, as they are major challenges for automated fact-checking. Active PETs is shown to be particularly beneficial with low labelling budgets and becomes less so when the labelling budget increases and/or the unlabelled pool is balanced. Furthermore, as Active PETs is built on PET, it inherits the limitations from PET, e.g. a pattern-verbaliser pair (PVP) is required for any classification tasks. Note that a good selection of tested PVPs that cover common NLP tasks are publicly available.

Our experiments are only conducted with PLMs that are of base and large sizes, e.g., BERT-base and BERT-large, due to limited computing resources. Future work may further experiment with giant models like T5-11b and GPT-3. Another interesting direction would be to extending the proposed voting mechanism such that giant models and tiny models can both contribute effectively in the same committee, e.g., GPT-3 and DistillBert. Ideally, despite that GPT-3 is much larger than DistillBert, the extended voting mechanism should still allow DistillBert to contribute effectively.



\bibliography{references}
\bibliographystyle{acl_natbib}

\appendix

\section{Algorithm Appendix}
\label{algo}

We present detailed pseudo code for proposed algorithms in this section. Algorithm \ref{query} executes a single query iteration with Active PETs. Algorithm \ref{training} executes the training loop with the option of conducting oversampling with Active PETs.

\begin{algorithm}
\caption{A Single Query Iteration}\label{query}
\begin{algorithmic}
\Require The last trained Commitee of PETs $C$, unlabelled data pool $U$, query size $k$

\Statex

\For{$PET_i \in C$}
\State $v_i \gets Size(PET_i)/\min_{\forall PET_i \in C} Size(PET_i)$ 
\EndFor
\Comment{assign number of votes}

\For{instance $x \in U$}
\For{$PET_i \in C$}

\State $V_{x_i} \gets resize(\hat{y}_{x_i}, v_i)$
\EndFor
\Comment{predict label and vote}

\State $S_x = -\sum_{\forall V_{x_i} \in V_x} \frac{V_{x_i}}{|V|} log\frac{(V_{x_i})}{|V|}$
\EndFor
\Comment{calculate entropy scores}
\end{algorithmic}

\Return $Sort(S)[:k]$ \Comment{return top k instances}
\end{algorithm}

\begin{algorithm}
\caption{Training}\label{training}
\begin{algorithmic}
\Require Labelled and sorted data $D$, A initial Commitee of PETs $C$

\Statex

\If{Oversampling}
    \State $c \gets max_{\forall class \in D} count(data \in class)$
    \State $D \gets resize_{\forall class \in D}(class, c)$
\EndIf \Comment{oversampling}

\For{$PET_i \in C$}
\State $PET_i \gets train(PET_i, D)$
\EndFor
\Comment{train the commitee of PETs}
\end{algorithmic}
\Return $C$ \Comment{return trained PETs}
\end{algorithm}

\section{Example Appendix}
\label{sec:appendix}
We present example instances from SCIFACT and cFEVER datasets in this section.
\begin{table*}[tb]
    \small
    \centering
    \begin{tabular}{p{3.5cm}p{10cm}p{1cm}}
        \toprule
        \multicolumn{3}{c}{\textbf{SCIFACT}} \\
        \midrule
        \multicolumn{1}{c}{\textbf{Claim}} & \multicolumn{1}{c}{\textbf{Evidence}} & \multicolumn{1}{c}{\textbf{Veracity}} \\
        \midrule
        ``Neutrophil extracellular trap (NET) antigens may contain the targeted autoantigens PR3 and MPO.'' &
        ``Netting neutrophils in autoimmune small-vessel vasculitis Small-vessel vasculitis (SVV) is a chronic autoinflammatory condition linked to antineutrophil cytoplasm autoantibodies (ANCAs). Here we show that chromatin fibers, so-called neutrophil extracellular traps (NETs), are released by ANCA-stimulated neutrophils and contain the targeted autoantigens proteinase-3 (PR3) and myeloperoxidase (MPO). Deposition of NETs in inflamed kidneys and circulating MPO-DNA complexes suggest that NET formation triggers vasculitis and promotes the autoimmune response against neutrophil components in individuals with SVV.'' &
        \textit{``Suppport''} \\
        \midrule
        ``Cytochrome c is transferred from cytosol to the mitochondrial intermembrane space during apoptosis.'' &
        ``At the gates of death. Apoptosis that proceeds via the mitochondrial pathway involves mitochondrial outer membrane permeabilization (MOMP), responsible for the release of cytochrome c and other proteins of the mitochondrial intermembrane space. This essential step is controlled and mediated by proteins of the Bcl-2 family. The proapoptotic proteins Bax and Bak are required for MOMP, while the antiapoptotic Bcl-2 proteins, including Bcl-2, Bcl-xL, Mcl-1, and others, prevent MOMP. Different proapoptotic BH3-only proteins act to interfere with the function of the antiapoptotic Bcl-2 members and\/or activate Bax and Bak. Here, we discuss an emerging view, proposed by Certo et al. in this issue of Cancer Cell, on how these interactions result in MOMP and apoptosis.'' &
        \textit{``Contradict''} \\
        \midrule
        ``Incidence of heart failure increased by 10\% in women since 1979.'' &
        ``Clinical epidemiology of heart failure. The aim of this paper is to review the clinical epidemiology of heart failure. The last paper comprehensively addressing the epidemiology of heart failure in Heart appeared in 2000. Despite an increase in manuscripts describing epidemiological aspects of heart failure since the 1990s, additional information is still needed, as indicated by various editorials.'' &
        \textit{``Neutral''} \\
        \bottomrule
        \toprule
        \multicolumn{3}{c}{\textbf{Climate FEVER}} \\
        \midrule
        \multicolumn{1}{c}{\textbf{Claim}} & \multicolumn{1}{c}{\textbf{Evidence}} & \multicolumn{1}{c}{\textbf{Veracity}} \\
        \midrule
        ``In 2015, among Americans, more than 50\% of adults had consumed alcoholic drink at some point.'' &
        ``For instance, in 2015, among Americans, 89\% of adults had consumed alcohol at some point, 70\% had drunk it in the last year, and 56\% in the last month.'' &
        \textit{``Suppport''} \\
        \midrule
        ``Dissociative identity disorder is known only in the United States of America.'' &
        ``DID is diagnosed more frequently in North America than in the rest of the world, and is diagnosed three to nine times more often in females than in males.'' &
        \textit{``Contradict''} \\
        \midrule
        ``Freckles induce neuromodulation.'' &
        ``Margarita Sharapova (born 15 April 1962) is a Russian novelist and short story writer whose tales often draw on her former experience as an animal trainer in a circus.'' &
        \textit{``Neutral''} \\
        \bottomrule
    \end{tabular}
    \caption{Veracity classification samples from the SCIFACT and Climate FEVER datasets.}
    \label{examples}
\end{table*} 

\section{Reproducibility Appendix}
\label{reproducibility}

We present further experimental details here for reproducibility. 
\paragraph{Number of parameters in each model} The number of parameters for BERT-base, BERT-large, RoBERTa-base, RoBERTa-large, DeBERTa-base, DeBERTa-large is 109484547, 335144963, 124647939, 355362819, 139194627 and 406215683 respectively.

\paragraph{Computing infrastructure} We use High Performance Compute cluster supported by the university. Each experiment is run with 8 compute cores, 11G RAM per core and a single NVIDIA A100 GPU.

\paragraph{Average run time} Table \ref{run_time} presents estimated average run time per sampling method per iteration using the computing infrastructure as described above. Specifically, we report the average run time of executing a sampling iteration of 150 unlabelled instances and a training iteration with the sampled data over three datasets. As CAL requires an initial labelled set of data, we report the total run time of an iteration of using the random method for 75 instances and an iteration of using CAL method for another 75 instances.
\begin{table}[htb]
    \footnotesize
    \centering
    \begin{tabular}{ccc}
        \toprule
         & \textbf{All Six Models} & \textbf{Average Single Model} \\
        \midrule
        \textbf{Random} & 00:05:50 & 00:00:58 \\
        \midrule
        \textbf{BADGE} &  00:07:52 & 00:01:19 \\
        \midrule
        \textbf{CAL} & 00:14:59 & 00:02:30 \\
        \midrule
        \textbf{ALPS} & 00:07:21 & 00:01:14 \\
        \midrule
        \textbf{Active PETs} & 00:08:01 & 00:01:20 \\
        \midrule
        \textbf{Active PETs-o} &  00:18:19 & 00:03:03 \\
        \midrule
        \bottomrule
    \end{tabular}
    \caption{Average run time for a single iteration for each of the sampling methods. The time format is hours:minutes:seconds.}
    \label{run_time}
\end{table}

\end{document}